\documentclass[11pt]{article}

\usepackage[preprint]{acl}

\usepackage{times}
\usepackage{latexsym}

\usepackage[T1]{fontenc}

\usepackage[utf8]{inputenc}

\usepackage{microtype}

\usepackage{inconsolata}

\usepackage{graphicx}
\usepackage{url}
\usepackage[greek,english]{babel}
\usepackage{enumitem}
\usepackage{multirow}

\usepackage{import}
\usepackage{amsmath}
\usepackage{amssymb}
\usepackage{tikz}
\usepackage{caption}
\usepackage{booktabs}
\usepackage{tabularx}
\usepackage[table]{xcolor} 
\usepackage{subcaption}
\usepackage{tikz}

\title{Proverbs or Pythian Oracles? Sentiments and Emotions in Greek Sayings}

\author{
 \textbf{Katerina Korre\textsuperscript{1}},
 \textbf{John Pavlopoulos\textsuperscript{1,2}},
\\
\\
\\
 \textsuperscript{1} Archimedes/Athena Research Center, Greece,\\
 \textsuperscript{2}Athens University of Economics and Business, Greece,
\\
\small{
  \href{mailto:k.korre@athenarc.gr}{k.korre@athenarc.gr},
  \href{mailto:annis@aueb.gr}{annis@aueb.gr}
}
}

\begin{document}
\maketitle
\begin{abstract}
Proverbs are among the most fascinating language phenomena
that transcend cultural and linguistic boundaries. Yet, much of the global landscape of proverbs remains underexplored, as many cultures preserve their traditional wisdom within their own communities due to the oral tradition of the phenomenon. Taking advantage of the current advances in Natural Language Processing (NLP), we focus on Greek proverbs, analyzing their sentiment and emotion. Departing from an annotated dataset of Greek proverbs, (1) we propose a multi-label annotation framework and dataset that captures the emotional variability of the proverbs, (2) we up-scale to local varieties \foreignlanguage{greek}{(ντοπιολαλιές)}, (3) we sketch a map of Greece that provides an overview of the distribution of emotions. Our findings show that the interpretation of proverbs is multidimensional, a property manifested through both multi-labeling and instance-level polarity. LLMs can capture and reproduce this complexity, and can therefore help us better understand the proverbial landscape of a place, as in the case of Greece, where surprise and anger compete and coexist within proverbs.

\end{abstract}


\section{Introduction}\label{sec:introduction}
Proverbs are a unique linguistic phenomenon that carries cultural significance, metaphorical wisdom, and timeless relevance. The latter characteristic can be seen as the symptom of the subjectivity of interpretation. \citet{martisolano:hal-03504178} discuss how relevance is timeless while meaning can change, with one of their use cases being \textit{a rolling stone gathers no moss}. While the proverb once urged people to remain in one place and build a steady future, with \textit{moss} symbolizing stability, the metaphoric \textit{moss} has more recently taken on a negative meaning, suggesting something that holds you back from exploring new opportunities. Studies like this are parts of two `sibling subfields' that are part of broader fields like linguistics, philology, and folkloristics, called \textit{paremiology} and \textit{paremiography}, and which are concerned with the study, collection and classification of proverbs, as well as addressing questions on their definition, form,  structure, style, content, function, meaning and value \cite{mieder2004proverbs}. 

The collaboration of paremiology (and/or paremiography) and Natural Language Processing (NLP) is still quite sparse. Most paremiology studies focus on a handful of use cases of proverbs with a very narrowed-down focus such as pragmatical or social aspects, as in the case of \citet{LOMOTEY202186}, while NLP studies in most cases utilize proverbs for their metaphorical aspect to help in broader tasks, such as metaphor detection or translation \cite{zbal2016, Romaniuk-Cholewska2024,  goren-strapparava-2024-context, wang-etal-2025-proverbs}. The limited overlap  is also driven by a major obstacle: proverbs, being rooted in oral tradition, are difficult to date, making it challenging to determine when they first got established \cite{Khanaifsawy2023}. This, in turn, complicates the task of compiling data for large-scale paremiological analysis. Another inhibiting factor is the `tautological’ tendency of technology—developing technology for its own sake—often at the expense of applying it to the digital humanities or to more traditional linguistic research.

Recognizing the importance of cultural heritage and the opportunities afforded by recent advances in NLP, we focus on the sentiment and emotion of Greek proverbs. Our motivation lies on two axes: First, on the idiosyncratic yet culturally and socially relevant nature of proverbs: although they may be structurally fixed, they are not semantically fixed. This semantic variability (and, consequently, variability in sentiment and emotion) indicates that meaning, sentiment, and emotion depend on time and location. As also noted by \citet{gwozdz2016sentiment}, ``proverbs are emotionally saturated \ldots\ and a close examination of the valency of paremic context may provide an insight into the usage of particular adages''. At the time that chapter was written, however, the author acknowledged that the state of AI was not advanced enough to achieve satisfactory accuracy. Here, we revisit the task in light of the progress brought by the large language models (LLMs), focusing on the variable of \textit{location}, as studying local  such differences helps us better understand the multidimensionality of proverbs as well as gaining insights into social values and cultural change.

The second axis is concerned with the fact that the proverbial landscape of Greece (similar to that of many medium- and low-resourced languages) remains largely understudied, particularly with regard to its local varieties \foreignlanguage{greek}{(ντοπιολαλιές}, ntopiolaliés),. Additionally, although one of the most definitive aspects of proverbs is their structural rigidity, paradoxically enough, they also appear in multiple variants, a characteristic that makes proverbs tough to detect and a challenge for corpus studies \cite{Norrick+2015+7+27, Kljajevic2025}.  To address these gaps, we provide a new dataset and analysis that not only serve as research bases for the study of the sentiment and emotion of proverbs but also establish a foundation for further research.


Our study is guided by one main research question (RQ$_m$) and three supporting sub-questions (RQ\textsubscript{s1--3}): 

\begin{enumerate}[label=\textit{RQ$_m$}:, leftmargin=2.5em]
  \item What is the sentimental and emotional interpretation of different proverbs in different Greek regions?
    \begin{enumerate}[label=\textit{RQ\textsubscript{s\arabic*}}:, leftmargin=3em]
        \item What differentiates proverbs where sentiment and emotion are aligned for most people from ones where they diverge, allowing for multiple possible affective interpretations?
        \item Can we aggregate and map the sentiment and emotion of proverbs to specific geographical locations?
        \item To what extent can LLMs capture and reflect the sentimental variance observed in different proverbs?
    \end{enumerate}
\end{enumerate}

Our findings show that (1) proverbs are multi-dimensional, and that conventional NLP approaches cannot fully capture the variability of their interpretation. Instead, we propose a pathway where both multi-label and polarity instances are taken into account for evaluation. (2) LLMs are able to capture multi-dimensionality but can over-predict some classes such as neutral for sentiment and angry and happy for emotion. (3) Aggregating data over large geographic areas (in our case with Greece) can lead to a significant loss of information by obscuring meaningful local variation. 

The rest of the paper is structured as follows. We begin by introducing related work (§\ref{sec:related_work}), covering NLP approaches utilizing and examining proverbs, as well as mentioning important works on proverbial sentiment and Greek proverbs. We then present our methodology (§\ref{sec:methodology}), which leads to the three main result sections: representation of proverb multidimensionality (§\ref{sec:agreement}), model experiments (§\ref{sec:model_experiments}), and error analysis (§\ref{sec:error_analysis}). These are followed by a discussion an expansion to a larger dataset (§\ref{sec:map}) and a (§\ref{sec:discussion}) in which we highlight key findings and their implications for paremiological analysis. We close with some remarks and outline directions for future work (§\ref{sec:conclusions}), limitations and ethical implications.

\section{Related Work}\label{sec:related_work}
Proverbs have been studied across disciplines, from folklorists who view them as carriers of folk wisdom and traditional lore, to linguists and lexicographers who analyze their linguistic features and document their usage \cite{Norrick+2015+7+27}. In this paper, we integrate these perspectives through computational methods to explore how different proverb interpretations express sentiment and emotion. This section is mainly informed by literature in NLP and computational linguistics.

\subsection{Proverbs in NLP}
Proverbs have been used to evaluate LLMs across diverse tasks in low-resource language settings. For example, \citet{azime-etal-2025-proverbeval} evaluated LLMs by prompting them both in the proverbs’ native languages and in English to assess their understanding of proverb meanings. The evaluation spanned multiple formats, including multiple-choice questions, fill-in-the-blank tasks, and a generation task in which models were required to produce the proverb that best matched a given description. The results showed that LLMs underperformed in the non-English settings. This gap in cultural and metaphorical understanding on the part of LLMs was also demonstrated in the work of \citet{Liu2024}, who introduced the MAPS dataset and evaluated various multilingal LLMs on them. Another proverb-based dataset is ePIC by \citet{ghosh-srivastava-2022-epic} who paired narrative and proverbs and showed that LLMs lack on proverb reasoning to a great degree compared to human annotators. With regard to dialects, \citet{magdy-etal-2025-jawaher} compiled a dataset of proverbs in Arabic varieties showing that although LLMs can produce relatively accurate translations, their cultural understanding is still limited. 

\subsection{Proverbial Sentiment}
Proverbs and sentiment analysis are often seen as complementary tasks, but with the main target being the improvement of sentiment analysis. One exception is \citet{ghosh-srivastava-2022-epic} who apart from narrative construction around proverbs, they also perform a small scale sentiment analysis to ensure the diversity of the narrative.  Comparatively, more work has been conducted on idioms and sentiment rather than proverbs and sentiment, as in the cases of \citet{WILLIAMS20157375}, \citet{jochim-etal-2018-slide}, \citet{tahayna2022}, and \citet{hwang-hidey-2019-confirming}. Despite all efforts, the sentiment and emotion of proverbs for cultural, linguistic, and ethnographic reasons remains much unexplored, with even less room left for non-English proverbs.

\subsection{What about Greek Proverbs?}
NLP studies on Greek proverbs are quite rare. \citet{Pavlopoulos2024} introduce a publicly-available dataset with more than one hundred thousand Greek proverb variants. They also proceed to attribute classification, showing that specific locations are classified
with higher accuracy than others, while also attributing more than 3k unregistered Greek proverbs using conformal prediction to estimate their most likely origins. At the low-resource language frontier, \citet{dimakis-etal-2025-dialect} use proverbs for dialect normalization using LLMs, showing that experiments (repeated) on normalised proverbs ignore deeper dialectal attributes.

\subsection{Why can't we treat proverbial sentiment and emotion as a conventional task?}\label{ssec:whymultilabel}
As mentioned in the introduction, a single proverb can span a very broad spectrum of sentiments and emotions, often encompassing both polarities (typically positive and negative). This stands in contrast to most conventional sentiment and emotion detection tasks, where ambiguity usually remains close to a single dominant label of a single polarity \cite{Wu2020}. In the case of a sentence with more than one polarity, it is usually attributed to different aspects of the sentence, like in the case of the review sentence ``the food quality is decent, but the price is very steep'', which contains a two aspect sentiment \cite{Tao2020}. This analysis cannot be applied to proverbs as when a proverb evokes both positive and negative sentiments (or a multitude of emotions), these are not necessarily tied to separable aspects, but instead emerge from the proverb’s overall interpretation, cultural grounding, and pragmatic context. As a result, attempting to decompose a proverb into aspect-level sentiments is often ill-defined.

\section{Method}\label{sec:methodology}

\begin{figure}
    \centering
    \includegraphics[width=1\linewidth]{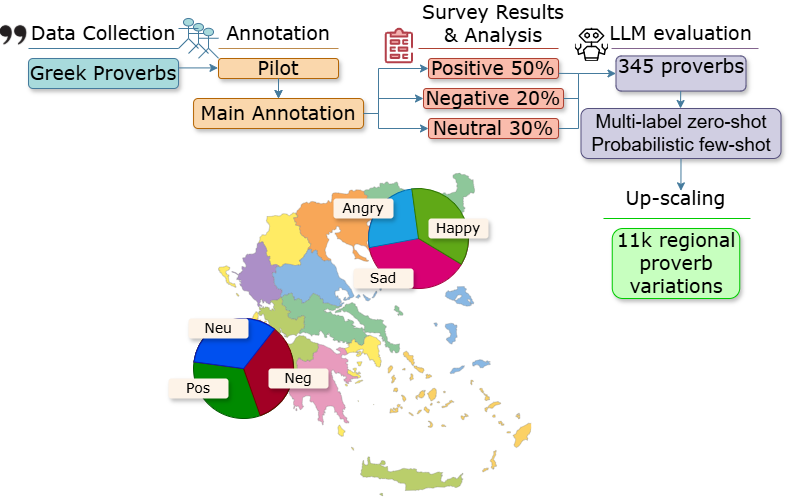}
    \caption{Methodology Pipeline.}
    \label{fig:pipeline}
\end{figure}

Proverbs may have multiple possible interpretations of proverbs, arising from personal, cultural, or contextual readings. These interpretations lead to mono- or multi-dimensional expressions of sentiment and emotion. We define \textit{monodimensional} proverbs as those that convey a dominant and relatively unambiguous sentiment or emotion, whereas \textit{multidimensional} proverbs evoke multiple, potentially coexisting or competing sentiments and emotions depending on the reader’s interpretation. 
Our work proposes a methodology for the prediction of monodimensional and multidimensional proverbs. We then move on to a larger scale regional application, analyzing the emotion of the proverbs per location. While this approach does not capture individual variations in the emotional perception of the proverbs, it provides a basis for a meaningful analysis of regional tendencies in social values and cultural expression.

Our methodology consists of five phases (see Figure~\ref{fig:pipeline}). The first phase concerns the collection of proverbs, while the second involves a multi-label annotation of sentiment and emotion. In the third phase, we analyze inter-annotator agreement and the distribution of labels as a more effective and honest way of representing the sentimental and emotional properties of the dataset. The fourth phase employs zero-shot and in-context learning techniques with LLMs to predict sentiment and emotion in our new dataset and to evaluate different types of prompting settings and models. The final phase involves an up-scaled inference setting for the geographical examination and visualization of the proverbial map of Greek varieties. 


\subsection{Dataset Development}\label{subsec:dataset_annotation}

\paragraph{Data Collection.}
We use two types of proverbs. For the first type, we use the term \textit{standard}, which corresponds to modern Greek without any recognised dialectal attributes.
These proverbs were collected from \textit{gnomikologikon}, an online collection of proverbs, sayings, and quotes.\footnote{\url{https://www.gnomikologikon.gr/}} The second type are proverbs with dialectal attributes or `locality', therefore we refer to these proverbs as \textit{localized} proverbs. An example of the two types can be found in Table~\ref{tab:dialect_example}. 
The localized proverbs were the ones compiled by \citet{Pavlopoulos2024}, which come from the Hellenic Folklore Research Centre.\footnote{\url{https://kentrolaografias.gr/}. The data are under a CC BY-NC-ND 4.0 licence.}

\begin{table}[t!]
\centering
\small
\setlength{\tabcolsep}{4pt} 
\renewcommand{\arraystretch}{1.2} 
\begin{tabular}{p{0.45\columnwidth}|p{0.45\columnwidth}}
\toprule
\textbf{Standard} & \textbf{Localized (Naxos Dialect)} \\
\hline
\textgreek{της νύχτας τη δουλειά τη βλέπει η μέρα και γελά} &
\textgreek{τση νύχτας τσοί δουλειές τσοί βλέπεις μέρα και γελάς} \\
\bottomrule
\end{tabular}
\caption{Example comparing a Standard Greek proverb with a Naxian dialect proverb. The word-by-word translation is \textit{the day sees the work of night and laughs}, meaning that whatever is done at night is done without much attention.}\label{tab:dialect_example}
\end{table}

\paragraph{Annotator Profiles.} We employ 12 full-time expert annotators that are native Greek speakers, possessing advanced knowledge in language technology and with prior annotation experience. The gender distribution was fairly balanced. 

\paragraph{Annotation Procedure.} 
We begin with a pilot annotation of 20 examples of Standard Greek proverbs. Annotators were asked to label sentiment (positive, negative, neutral) and emotion, using the outer ring of the \citet{Willcox1982} `wheel of emotions'. This allows us to proceed from a finer- to a coarser-grained analysis by later mapping emotions from the outer to the inner ring. You can find the wheel in Appendix~\ref{app:sec:guidelines_and_wheel}, Figure~\ref{app:fig:emotions}. The annotation setup is multi-label (\S\ref{ssec:whymultilabel}), allowing more than one sentiment or emotion to be assigned to a single proverb. In the case of more than one sentiment/emotion, the annotators had to use a separate field by justifying their answer and providing the corresponding interpretations. Moreover, in cases where an emotion was absent from the list, annotators were instructed to select the closest available emotion and then use an additional field to specify an alternative emotion they considered as more appropriate. Annotators provided also a confidence score (see Appendix~\ref{app:sec:confidence}) and optional comments. 

Inter-annotator agreement was low (see Appendix~\ref{app:sec:guidelines_and_wheel}, Table~\ref{app:tab:alpha_pilot}), although the annotators justified fairly their decisions, leaving no room for resolution. In other words, a single proverb could have multiple emotions (or sentiment) labels, depending on the annotator. Hence, we keep the same settings for the main annotation round. 
The main annotation round included 345 proverbs. We selected unique proverbs from each region (i.e., ones existing in one region only) to capture region-specific sentiment and emotional expression. In addition to the pilot setup, and given the dialectal nature of our dataset, we provided a normalized version of each proverb in Standard Greek. The quality of the normalization was reviewed by the authors of the paper, who made any necessary modifications. Moreover, we added an extra field to the annotation document, allowing annotators to flag cases of inadequate normalization. The results of the analysis of the agreement on both of these aspects is discussed in \S\ref{sec:agreement}.



\subsection{Model Evaluation and Experiments.}
\paragraph{Evaluation on Localized Proverbs} We evaluate LLMs on the annotated dataset using the same multilabel sentiment and emotion schema as in human annotation through prompting. The prompts can be found in Tables~\ref{app:tab:prompts_sentiment} and~\ref{app:tab:prompts_emotion} in Appendix~\ref{app:sec:prompts}.\footnote{For the emotions, we used directly the inner ring list of emotions as early experiments showed that the models have to choose from too many labels.} Model predictions are compared against the gold labels using macro-averaged F1 scores, computed per emotion and sentiment category and aggregated by region. In addition to overall performance, we analyze the number of emotions and sentiments predicted per proverb and the probability for each sentiment or emotion.

\paragraph{Regional Application.} To study model behavior beyond the limits of manual annotation, we conduct a second, exploratory phase using approximately 11k unannotated proverb instances. These instances are automatically labeled by the same LLMs using in-context learning methods. We explicitly do not treat these automatically generated labels as ground truth but as a way to assess whether large-scale automatic annotation preserves the properties observed in the gold dataset. 


\paragraph{In-Context Learning Methods.}
We performed zero-shot and few-shot learning, using both open-source and close-source models, including two specifically designed for Greek language tasks.
In an attempt to have a variety with regard to models, we chose relatively small LLMs (Krikri-8B and Mistral-7B) as well as two larger ones (GPT-5mini and Llama-70B). We included variations of zero-shot settings, summarized as follows:  

\begin{itemize}

  \item \textbf{Simple Zero-shot (\(Z_0\)):}  
  Given an input instance \(x\), the model is prompted to directly predict the label 
  \(\hat{y}\) without being provided with any labeled examples.

  \item \textbf{Probabilistic Prediction Prompting (\(Z_p\)):}  
  Instead of predicting a single label per text $x$, the model is instructed to output a probability 
  distribution \(\hat{p}(y \mid x)\) over all sentiment and emotion classes $y$.  
  Each class is provided with %
  examples, where 
  the examples include percentage-based annotations designed to cover all possible 
  sentiments and emotions by explicitly varying their intensity.  
  This formulation captures model uncertainty and allows for more flexible evaluation.

\end{itemize}

  


\section{Results}\label{sec:results}

\subsection{Not disagreement but multidimensionality}\label{sec:agreement}

\begin{table}[t!]
\centering
\small
\begin{tabular}{p{4cm}ccc}
\hline
\textbf{Group} & \textbf{Pos} & \textbf{Neg} & \textbf{Neu} \\
\hline
\textbf{Semantic Grouping} & & & \\
Pos + Neutral (pos\_neu) & 0.383 & - & - \\ 
Neg + Neutral (neg\_neu) & - & 0.370 & - \\
Neutral only & - & - & 0.175 \\
\hline
\textbf{Variance Types} & & & \\
Polarity disagreement & 0.091 & 0.151 & - \\
Multilabel conflict & 0.367 & 0.122 & - \\
\hline
\textbf{Unimodal Items} & 0.671 & 0.466 & - \\
\hline
\textbf{Overall $\alpha$ per Label} & 0.383 & 0.362 & 0.192 \\
\textbf{Global $\alpha$} & \multicolumn{3}{c}{0.312} \\
\hline
\end{tabular}
\caption{Krippendorff's $\alpha$ for Sentiment Annotations: Semantic Groups, Item Types, and Overall}
\label{tab:krippendorff_sentiment}
\end{table}

We analyze the dimensionality of the annotations for sentiment in three different ways. The results are shown in Table~\ref{tab:krippendorff_sentiment}. We use this method as a diagnostic to characterize and better analyze types of ambiguity in proverbs (\S\ref{ssec:whymultilabel}). In addition to the overall Krippendorff’s $\alpha$ scores, we group the annotations into semantic clusters (pos\_neu, neg\_neu, and neutral), where neutral labels can also contribute to the two polarities. Agreement was moderate for the polarity-related groups, with pos\_neu reaching $\alpha = 0.38$ and neg\_neu $\alpha = 0.37$. In contrast, strictly neutral annotations showed lower agreement ($\alpha = 0.17$), suggesting that neutral labels were applied more inconsistently and may primarily reflect annotator uncertainty.

We further group annotations by variance type, because multidimensionality can arise either from \textbf{polarization}, when some annotators label the same instance as positive while others label it as negative; or from \textbf{multilabeling}, when annotators assign both positive and negative labels to the same instance. As expected, we observe very low agreement for polarizing items, where annotators selected opposite single-label polarities, with $\alpha = 0.09$ for positive and $\alpha = 0.15$ for negative labels. Items with multiple labels show somewhat higher agreement for positive labels ($\alpha = 0.37$) but still low agreement for negative labels ($\alpha = 0.12$). Finally, unimodal items, where annotators consistently agree on a single polarity, exhibit the highest agreement, with $\alpha = 0.67$ for positive labels and $\alpha = 0.47$ for negative labels. Example cases of the types of variance can be found in Table~\ref{tab:example_types}.

\begin{table}[t!]
\centering
\small
\begin{tabular}{p{3cm} c}
\hline
\textbf{Emotion} & \textbf{Krippendorff's $\alpha$} \\
\hline
happy       & 0.326 \\
angry       & 0.127 \\
bad         & 0.105 \\
fearful     & 0.103 \\
sad         & 0.091 \\
disgusted   & 0.079 \\
surprised   & 0.074 \\
\hline
\end{tabular}
\caption{Krippendorff's $\alpha$ for inner-ring mapped emotions.}
\label{tab:kripp_alpha_emotion}
\end{table}

\begin{table}[!t]
\centering
\small
\begin{tabular}{p{0.07\columnwidth}|p{0.4\columnwidth}|p{0.4\columnwidth}}
\toprule
\textbf{Type} & \textbf{Greek Proverb} & \textbf{English Translation} \\
\hline
Uni & \foreignlanguage{greek}{Θηβαίοι κι Αθηναίοι και κακοί Μυτιληναίοι άλλα λένε το βράδι κι άλλα κάνουν το πρωΐ} 
& Thebans and Athenians and wicked Mytileneans say one thing in the evening and do another in the morning. \\
PD & \foreignlanguage{greek}{Τσείνες π' μ' αγαπά, με κάνει τσαι κλαίου, τσαι τσείνες π' με μ'σά, με κάν' τσαι γελού} 
& The one who loves me makes me cry and the one who hates me makes me laugh. \\
MC & \foreignlanguage{greek}{Δούλευε και δουλειά να μη σε λείπ'} 
& Work hard and don't miss work.\\
\bottomrule
\end{tabular}
\caption{Examples of different types of annotator agreement: unimodal (Uni), polarity disagreement (PD), and multilabel conflict (MC).}
\label{tab:example_types}
\vspace{-10pt}
\end{table}

\begin{figure}
    \centering
    \includegraphics[width=0.8\linewidth]{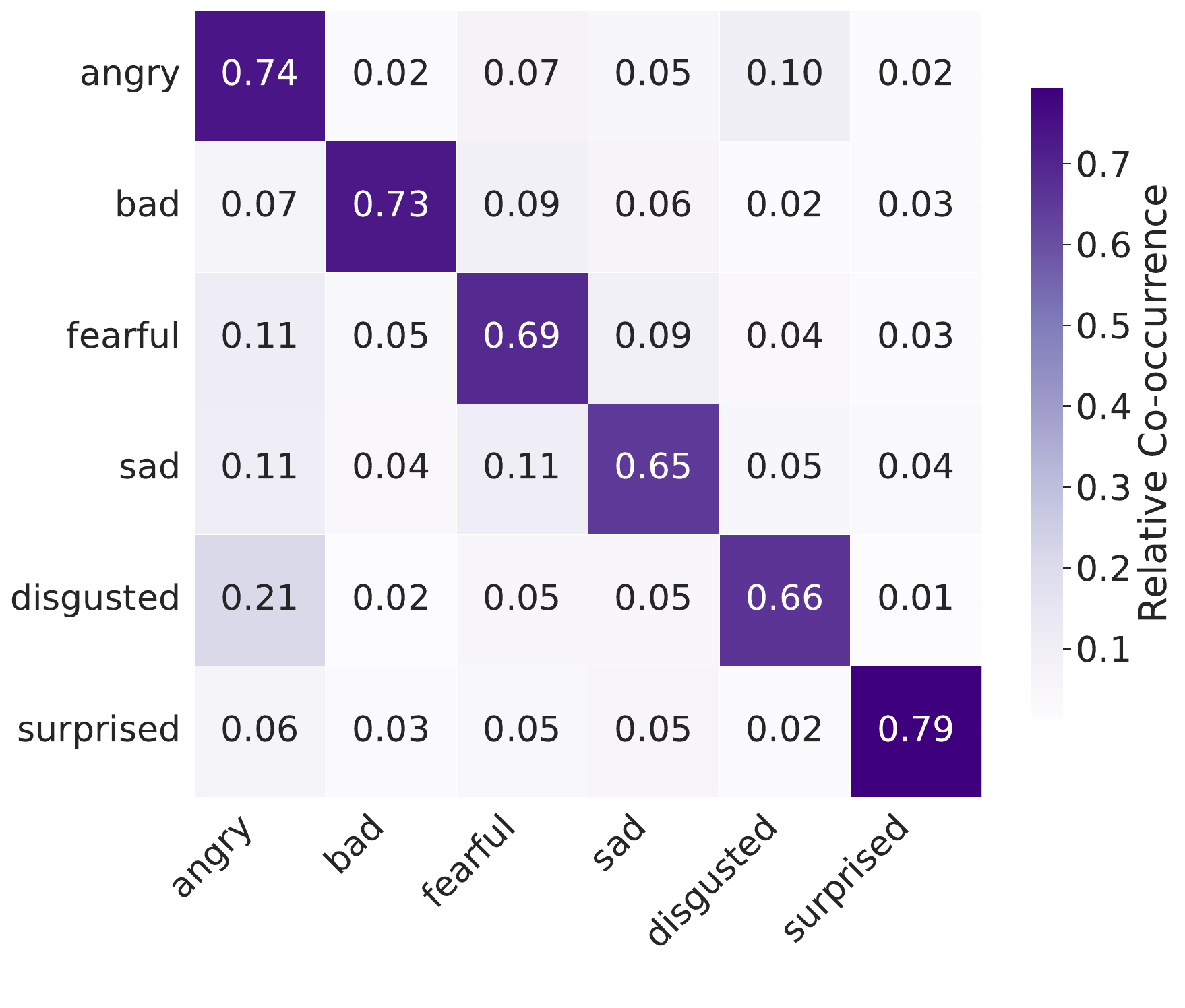}
    \caption{Relative co-occurence in annotations for low-agreement emotions.}
    \label{fig:cooccurrence_heatmap}
\end{figure}

To examine the multidimensionality of proverbs with regard to emotions, we first map the fine-grained emotions into the five more coarse-grained emotions in the \citet{Willcox1982} wheel of emotions. The agreement for each emotion can be found in Table~\ref{tab:kripp_alpha_emotion}. With a first glance, we see that higher agreement is aggregated in the positive sentiment of happiness, which reminds the results for the sentiment analysis as well. For the rest of the sentiments that showed lower agreement, we wanted to see each low-agreement emotion with which emotion co-exists. We define a low-agreement threshold of 0.2, which will include all emotions except happy. For each low-agreement emotion, we (1) count how often it co-occurs with every other emotion in the same proverb, (2) normalize by row sums so each row sums to 1 (proportion of co-occurrences). The results can be found in Figure~\ref{fig:cooccurrence_heatmap}. Each row represents a reference emotion, and the values indicate the proportion of proverbs in which other emotions co-occur with it.  Angry presents minor overlaps with fearful (0.11) and sad (0.11), while sad co-occurs with with fearful and angry (both 0.11). Disgusted shows a co-occurrence with angry (0.21) and surprised largely appears independently (0.79), with minimal co-occurrence with other emotions. These patterns show that disagreements among annotators often reflect true multi-label characteristics of the dataset: \emph{many proverbs express multiple emotions simultaneously}.

These results are also reflected in the examples of Table~\ref{tab:example_types}. For the unimodal/monodimensional proverb, the majority of annotators agreed it conveyed a negative sentiment, with fine-grained emotions such as betrayal and resentment identified. 

The example with polarity disagreement was labeled with all three labels (7 neutral, 2 positive, 3 negative). Since this was not a multilabeled instance, there were not any interpretations by the annotators, but we can use the emotions to better understand the root of the disagreement. The negative sentiment was supported by emotions such as perplexity, disillusionment, and hostility, while the positive sentiment was interpreted through emotions such as loveliness and sensitivity. The neutral instances included justifications referencing all the emotions found in the two polarities. This can be seen as a result of the two-part structure of the proverb, which reflects a known challenge in sentiment analysis: concepts that contradict each other (e.g., love and hate, joy and sorrow) can co-exist within a single sentence adding to the multi-dimensionality of the proverb.
The multilabel conflict example shows a more personal interpretation. While the majority of annotators labeled the instances as positive or neutral, with emotions such as inspired, energetic, hopeful, and thankful, there were also instances where it was viewed as negative, with the emotion pressured. This highlights the very intimate relationship between a proverb and its receiver.

\begin{table*}[t!]
\centering
\small
\begin{tabular}{l l 
                c c c c c c 
                c c c c c c}
\toprule
\multirow{2}{*}{\textbf{Model}} & \multirow{2}{*}{\textbf{Setting}} 
& \multicolumn{6}{c}{\textbf{Sentiment}} 
& \multicolumn{6}{c}{\textbf{Emotion}} \\
\cline{3-14}
 & & \multicolumn{3}{c}{Micro} & \multicolumn{3}{c}{Macro} 
 & \multicolumn{3}{c}{Micro} & \multicolumn{3}{c}{Macro} \\
\cline{3-14}
 & & P & R & F1 & P & R & F1 & P & R & F1 & P & R & F1 \\
\hline
KriKri-8B  & Zero-shot & 0.61 & 0.53 & 0.57 & 0.48 & 0.53 & 0.47 & 0.28 & 0.63 & 0.39 & 0.33 & 0.65 & 0.41 \\
Mistral-7B  & Zero-shot & 0.64 & 0.61 & 0.62 & 0.65 & 0.62 & 0.60 & 0.27 & 0.38 & 0.32 & 0.27 & 0.39 & 0.31\\
GPT-5mini & Zero-shot & 0.78 & 0.74 & 0.76 & 0.79 & 0.77 & 0.75 & 0.23 & 0.38 & 0.28 & 0.22 & 0.40 & 0.27 \\
Llama3-70B & Zero-shot & 0.61 & 0.75 & 0.67 & 0.61 & 0.78 & 0.66 & 0.25 & 0.46 & 0.32 & 0.24 & 0.47 & 0.31 \\
\midrule
KriKri-8B  & Probabilistic & 0.79 & 0.52 & 0.62 & 0.80 & 0.57 & 0.64 & 0.34 & 0.27 & 0.30 & 0.34 & 0.28 & 0.30 \\
Mistral-7B  & Probabilistic & 0.76 & 0.54 & 0.63 & 0.76 & 0.57 & 0.62 & 0.22 & 0.20 & 0.21 & 0.21 & 0.20 & 0.20\\
GPT-5mini & Probabilistic & 0.74 & 0.73 & 0.74 & 0.76 & 0.77 & 0.73 & 0.54 & 0.40 & 0.46 & 0.54 & 0.43 & 0.47 \\
Llama3-70B & Probabilistic & 0.72 & 0.63 & 0.67 & 0.76 & 0.67 & 0.68 & 0.44 & 0.41 & 0.42 & 0.47 & 0.45 & 0.44 \\\hline
Human-baseline & & &  & 0.62 & &  & 0.59 &  &   & 0.25 & & 0.18\\
\bottomrule
\end{tabular}
\caption{Evaluation of multilabel emotion/sentiment classification of Greek proverbs with different models/settings.}\label{tab:emotion_sentiment_models}
\end{table*}


\subsection{LLM-based Evaluation}\label{sec:model_experiments}
We use the 345 annotated standard proverbs to examine the potential of LLMs to capturing the multi-dimensional nature of proverbs. Polarized instances were modeled as multi-label: categories with a score > 0.3 were assigned, allowing an instance to belong to multiple categories simultaneously. For the probabilistic few-shot evaluation, predicted sentiment and emotion percentages are converted to a 0–1 scale and transformed into discrete predicted labels using a probability threshold, with a fallback to the highest-scoring sentiment or emotion if none exceed the threshold. Gold and predicted labels are then aligned per proverb, encoded as multi-hot vectors. All the results are presented in Table~\ref{tab:emotion_sentiment_models}.
A human-baseline, where each annotator's labels are compared to the majority of the rest by means of (mean) F1 score, quantifies the task difficulty, showing that the classification of emotion is more challenging than that of sentiment. Interestingly, LLMs perform equally or better than this baseline. For sentiment classification, GPT-5mini generally achieves the highest F1 scores in both zero-shot (micro F1 = 0.76) and probabilistic settings (micro F1 = 0.74). The main observation between the two different settings is that the probabilistic method improves the recall for some models but not consistently for all of them. For emotion classification, the scores are overall lower than sentiment classification, with Krikri-8B achieving the highest scores. We also observe that the probabilistic settings improve performance significantly for GPT-5mini for micro F1 score and Llama3-70B for macro F1. Overall, the probabilistic approach can be more meaningful for emotion prediction.

\subsubsection{Error Analysis}\label{sec:error_analysis}
We also perform an error analysis on the best-performing models (GPT-5mini and Krikri-8B) examining the co-occurrences of the labels. The metrics per label and Mean Squared Error can be found in Appendix~\ref{app:sec:error_analysis}.
With regard to sentiment zero-shot classification (see Table~\ref{tab:sentiment_error}) the Positive label is the rarest, and it is often under-predicted (confused with co-occurring labels). For the probabilistic setting, Neutral is overpredicted, as we see that many negatives and positives are misclassified as neutral. Positive, on the other hand is underpredicted again. Negative is predicted relatively better, but still often confused with neutral.

\begin{table}[t!]
\centering
\footnotesize
\setlength{\tabcolsep}{5pt}
\renewcommand{\arraystretch}{1.25}

\begin{subtable}{0.49\linewidth}
\centering
\begin{tabular}{c|ccc}
 & \textbf{neg} & \textbf{neu} & \textbf{pos} \\
\midrule
\textbf{neg} & \cellcolor{red!45!white}235 & \cellcolor{red!25!white}120 & \cellcolor{red!5!white}25 \\
\textbf{neu} & \cellcolor{red!25!white}120 & \cellcolor{red!42!white}222 & \cellcolor{red!20!white}52 \\
\textbf{pos} & \cellcolor{red!5!white}25 & \cellcolor{red!20!white}52 & \cellcolor{red!33!white}78 \\
\end{tabular}
\subcaption{Zero-shot}
\end{subtable}
\hfill
\begin{subtable}{0.49\linewidth}
\centering
\begin{tabular}{c|ccc}
 & \textbf{neg} & \textbf{neu} & \textbf{pos} \\
\midrule
\textbf{neg} & \cellcolor{blue!45!white}190 & \cellcolor{blue!35!white}172 & \cellcolor{blue!15!white}44 \\
\textbf{neu} & \cellcolor{blue!25!white}108 & \cellcolor{blue!40!white}162 & \cellcolor{blue!20!white}78 \\
\textbf{pos} & \cellcolor{blue!10!white}18 & \cellcolor{blue!30!white}68 & \cellcolor{blue!35!white}57 \\
\end{tabular}
\subcaption{Probabilistic}
\end{subtable}

\caption{Confusion matrices on sentiment (a) zero-shot and (b) probabilistic few-shot classification.}
\label{tab:sentiment_error}
\vspace{-10pt}
\end{table}

\begin{figure}[t]
\centering

\begin{subfigure}{\columnwidth}
\centering
\renewcommand{\arraystretch}{1.15}
\resizebox{\columnwidth}{!}{
\begin{tabular}{l|ccccccc}
\textbf{Gold / Pred} & angry & bad & disgusted & fearful & happy & sad & surprised \\
\hline
angry      & \cellcolor{red!60}98 & \cellcolor{red!40}60 & \cellcolor{red!30}17 & \cellcolor{red!30}18 & \cellcolor{red!70}110 & \cellcolor{red!55}81 & \cellcolor{red!30}18 \\
bad        & \cellcolor{red!25}15 & \cellcolor{red!20}11 & \cellcolor{red!15}3  & \cellcolor{red!15}3  & \cellcolor{red!35}22  & \cellcolor{red!30}17 & \cellcolor{red!10}2  \\
disgusted & \cellcolor{red!25}12 & \cellcolor{red!20}6  & \cellcolor{red!15}2  & \cellcolor{red!10}1  & \cellcolor{red!30}17  & \cellcolor{red!25}11 & \cellcolor{red!15}3  \\
fearful   & \cellcolor{red!40}36 & \cellcolor{red!35}25 & \cellcolor{red!30}11 & \cellcolor{red!30}10 & \cellcolor{red!40}37  & \cellcolor{red!40}30 & \cellcolor{red!20}6  \\
happy     & \cellcolor{red!40}36 & \cellcolor{red!35}26 & \cellcolor{red!30}11 & \cellcolor{red!25}9  & \cellcolor{red!60}103 & \cellcolor{red!50}71 & \cellcolor{red!35}20 \\
sad       & \cellcolor{red!30}16 & \cellcolor{red!30}16 & \cellcolor{red!25}8  & \cellcolor{red!25}7  & \cellcolor{red!30}20  & \cellcolor{red!35}27 & \cellcolor{red!10}1  \\
surprised & \cellcolor{red!30}14 & \cellcolor{red!25}7  & \cellcolor{red!20}2  & \cellcolor{red!20}2  & \cellcolor{red!35}29  & \cellcolor{red!30}16 & \cellcolor{red!20}5  \\
\end{tabular}} \subcaption{Zero-shot}
\end{subfigure}

\vspace{0.5em}

\begin{subfigure}{\columnwidth}
\centering

\renewcommand{\arraystretch}{1.15}
\resizebox{\columnwidth}{!}{
\begin{tabular}{l|ccccccc}
\textbf{Gold / Pred} & happy & angry & fearful & sad & disgusted & surprised & bad \\
\hline
happy      & \cellcolor{blue!35}36 & \cellcolor{blue!25}14 & \cellcolor{blue!10}1 & \cellcolor{blue!10}1 & 0 & \cellcolor{blue!60}65 & \cellcolor{blue!15}2 \\
angry      & \cellcolor{blue!30}20 & \cellcolor{blue!45}50 & 0 & \cellcolor{blue!10}1 & 0 & \cellcolor{blue!65}73 & \cellcolor{blue!20}3 \\
fearful    & \cellcolor{blue!20}5  & \cellcolor{blue!30}16 & \cellcolor{blue!10}1 & 0 & 0 & \cellcolor{blue!40}25 & \cellcolor{blue!15}2 \\
sad        & \cellcolor{blue!20}6  & \cellcolor{blue!25}12 & 0 & \cellcolor{blue!15}2 & 0 & \cellcolor{blue!25}11 & 0 \\
disgusted & \cellcolor{blue!15}3  & \cellcolor{blue!25}9  & 0 & 0 & 0 & \cellcolor{blue!25}9  & \cellcolor{blue!15}2 \\
surprised & \cellcolor{blue!20}8  & \cellcolor{blue!20}4  & 0 & 0 & 0 & \cellcolor{blue!40}20 & 0 \\
bad        & \cellcolor{blue!15}4  & \cellcolor{blue!20}5  & 0 & 0 & 0 & \cellcolor{blue!40}19 & 0 \\
\end{tabular}}\subcaption{Probabilistic}
\end{subfigure}

\caption{Emotion confusion matrices on (a) multilabel zero-shot, (b) probabilistic few-shot classification.}
\vspace{-10pt}
\end{figure}

For emotion, Happy is often misclassified as angry and vice versa (110). This suggests that the model struggles to differentiate positive vs negative emotions. Many fearful examples are predicted as happy (37), angry (36), sad (30), or bad (25). Sad is often misclassified as happy (20), angry (16) or bad (16). This error analysis reflects the fact that models can capture the mutlidimensionality of proverbs and that even polar opposites can exist within the same proverb. The overall pattern shows that the model tends to overpredict happy and angry, often at the expense of fearful, disgusted, and sad.

\section{Emotional Greek Proverbial Landscape }\label{sec:map}
\begin{figure}[!t]
    \centering
    \includegraphics[width=0.7\linewidth]{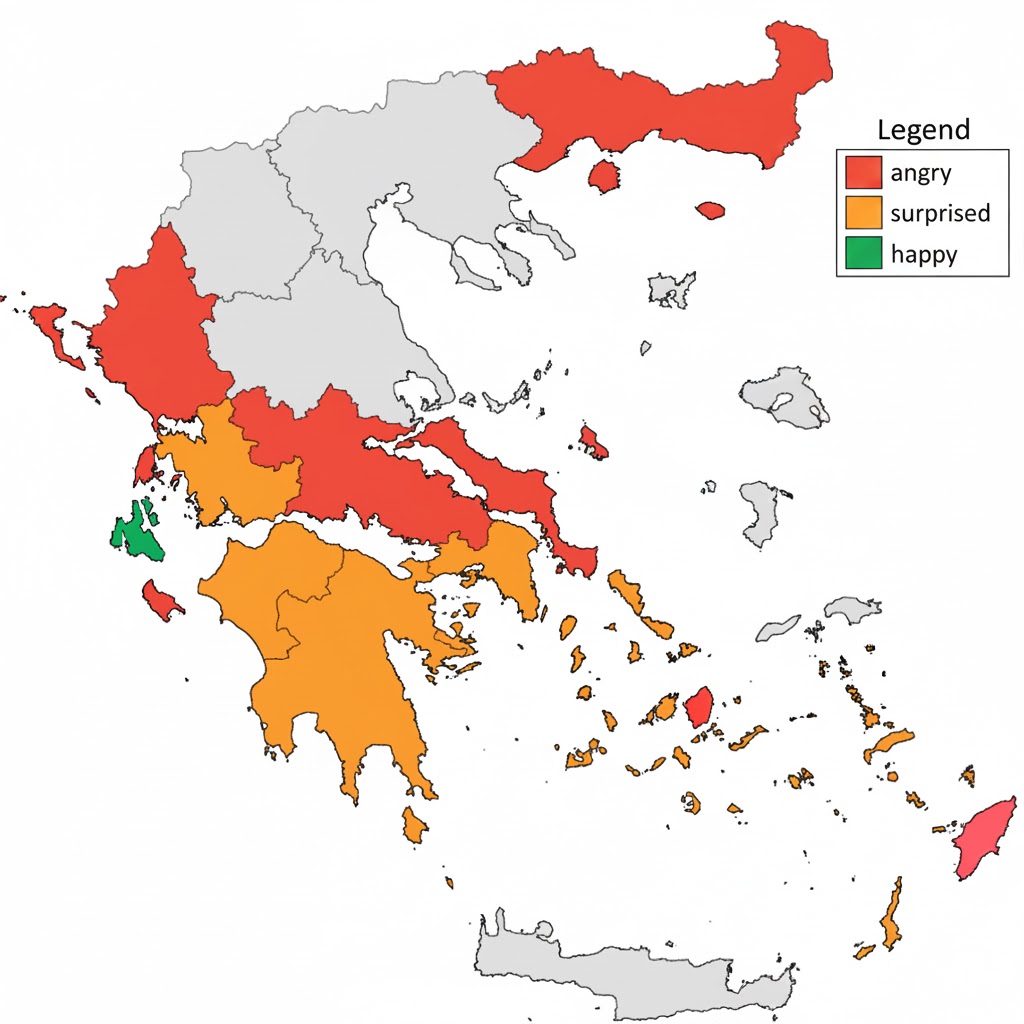}
    \caption{Emotion map of Greek proverbial landscape. Different colors represent the most dominant emotion. In gray are regions without any geolocated proverbs.}
    \label{fig:map}
\end{figure}

\begin{figure}[!t]
    \centering
    \includegraphics[width=1\linewidth]{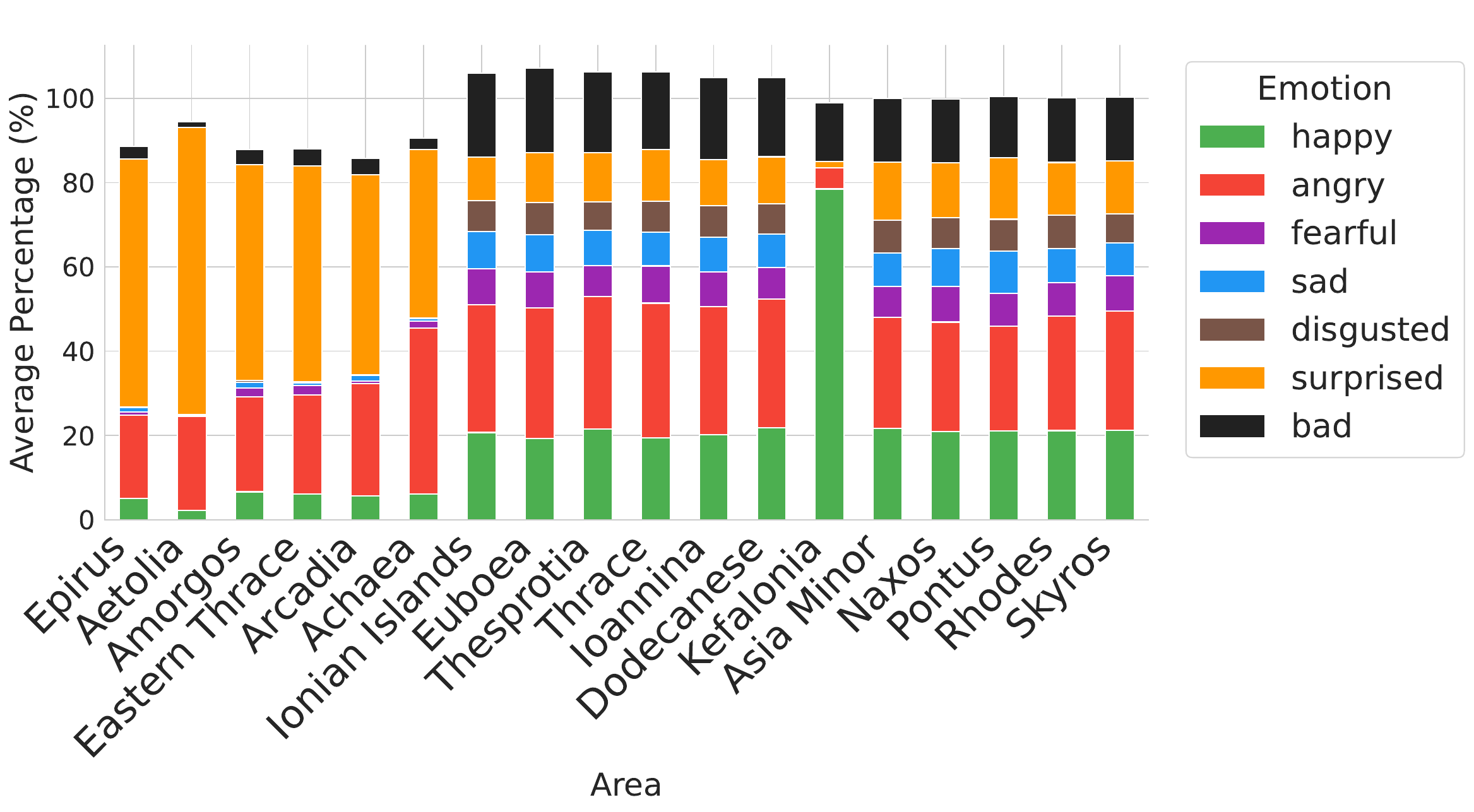}
    \caption{Proverbial emotions per major Greek areas.}
    \label{fig:emotion_distributions}
    \vspace{-10pt}
\end{figure}

Using Krikri-8B, the best in emotion classification--and better than the human baseline--model (see Section~\ref{sec:results}),\footnote{We use the same prompts as in our model experiments.} we up-scale on the dataset of 11k geographically attributed proverbs by \citet{Pavlopoulos2024}. We then visualize our findings on a proverbial map of Greece (Figure~\ref{fig:map}), illustrating the dominant emotions. At the same time, we capture the multidimensional nature of proverbs by providing the distribution of emotions across regions (Figure~\ref{fig:emotion_distributions}).
We observe that the two dominant emotions are anger and surprise, which effectively split the country in two, with very few exceptions (Kefalonia, Rhodes, Naxos). These exceptions, however, carry significant meaning because, as we see in the map, aggregating emotions at larger spatial scales causes local emotional variations to disappear, as in the case of Kefalonia where the emotion is `happy' while the emotion for the greater area of Ionian Islands is `anger'. The distributions also show that these two emotions, along with happiness, coexist in most regions, while the remaining emotions play a less prominent role.

\section{Discussion}\label{sec:discussion}
\paragraph{On the Research Questions} The sentiment and emotion of proverbs are neither homogeneous nor mono-dimensional. A single proverb can be interpreted differently in terms of sentiment, which may manifest both as annotators assigning multiple labels or as variations in polarity. These `conflicts' usually stem from personal interpretations, which challenge the notion of proverbs as fixed entities and demonstrate how, over time, individual interpretations can gradually become consolidated at a social level. 
Regionally, differences can appear on a large scale, as shown in Figure~\ref{fig:map}, where the landscape seems split in two. However, a closer look reveals more subtle variations. For example, in the Ionian Islands, the dominant sentiment is anger, except on the island of Kefalonia. Simple averaging across the region would obscure important information about this area.


\paragraph{On the Origin of Greek Proverbs} An analysis such as the one presented in this paper cannot give us any information  about the genesis of a specific proverb, but rather its spreading. An attempt on the origin of any proverb is futile, as described in much of the paremiological literature \cite{Norrick+2015+7+27}. This is because it is most likely that proverbs are coined by a single individual, who managed to grasp a universal truth in a metaphorical yet adequately expressive manner that it was adopted and used by the folk. This however and as discussed in previous sections implicates that the way the proverbs are used has led to variation. Here, although we map the proverbs and their emotion to specific regions in Greece, the only indication that we have that these proverbs `belong to' the specific regions are the dialectal properties of the proverbs. This however does not mean that one region `owns' a proverb, but rather it owns its specific variety. 


\section{Conclusions}\label{sec:conclusions}
In this paper, we employed NLP methods to address the task of sentiment and emotion analysis in Greek proverbs. Our approach examines the multidimensionality of proverbs, despite their seemingly `fixed' nature. To this end, we propose treating sentiment and emotion analysis not only as a multi-label task, but also explicitly accounting for polarity, and adapting LLM-based pipelines to support this more inclusive aspect. Finally, when conducting region-based analyses, fine-grained geographic units should not be overlooked, as they may capture important local variation.
Future work will take a deeper look at the probabilistic approach and examine threshold sensitivity to achieve optimal results. In addition, a more geographically fine-grained analysis of proverbs, combined with larger-scale annotation enriched with additional metadata, would allow for a more complete picture.

\section*{Limitations}\label{sec:limitations}
This study is not without limitations. The presence of the same proverbs across different localities does not mean necessarily that sentiment and emotion are perceived in the same way. On the contrary, it suggests that sentiment and emotion may vary depending on the place, the background and the experiences of the peoples. This highlights, once again, the inherently multidimensional nature of interpreting proverbs. Additionally, while some proverbs can be considered `gold' and be very similar in all the different varieties, others can be similar in meaning but different in form; yet both types reflect the oracle-like nature of these sayings. Their evaluation, however, is necessarily filtered through our own subjectivity, since no objective truth can be applied in this context. Another limitation concerns the geolocation of proverbs, because the 11K proverbs that are currently geolocated by \citet{Pavlopoulos2024}, leave areas with out any proverb for analysis (the gray regions in Figure~\ref{fig:emotion_distributions}). Future research however, could expand their list of geolocated proverbs, to include more areas. It is also important to note that we do not know the extent to which these proverbs are actively used today, leaving their current relevance somewhat uncertain. Finally, during groundtruth, polarity conflicts are treated here as a multi-label phenomenon, although in our methodology we note that they ideally should be handled differently. Current NLP studies rarely account for this type of subtle variation across categories, and while detection has been addressed individually, integrating it into evaluation remains an open challenge.

\section*{Ethics Statement}\label{sec:ethics}
Participation was compensated adequately in accordance with applicable Greek legislation. All proverbs used in this study were collected from publicly available sources and existing corpora. No private or sensitive data was used. We acknowledge that annotators’ interpretations of sentiment in  proverbs may reflect cultural or regional biases. This research is intended for academic purposes and cultural understanding, therefore, we acknowledge that automated sentiment and emotion predictions should be used carefully and with respect to the local (proverbial communities) and definitely not in order to stereotype one community as positive and another as negative.

\section*{Acknowledgments}\label{sec:acknowledgments}
This work has been partially supported by project
MIS 5154714 of the National Recovery and Resilience Plan Greece 2.0 funded by the European
Union under the NextGenerationEU Program.

\bibliography{acl_latex}

\clearpage

\appendix

\label{sec:appendix}
\section{Annotation Guidelines and Pilot Results}\label{app:sec:guidelines_and_wheel}
\begin{figure}[!t]
    \centering
    \includegraphics[width=1\linewidth]{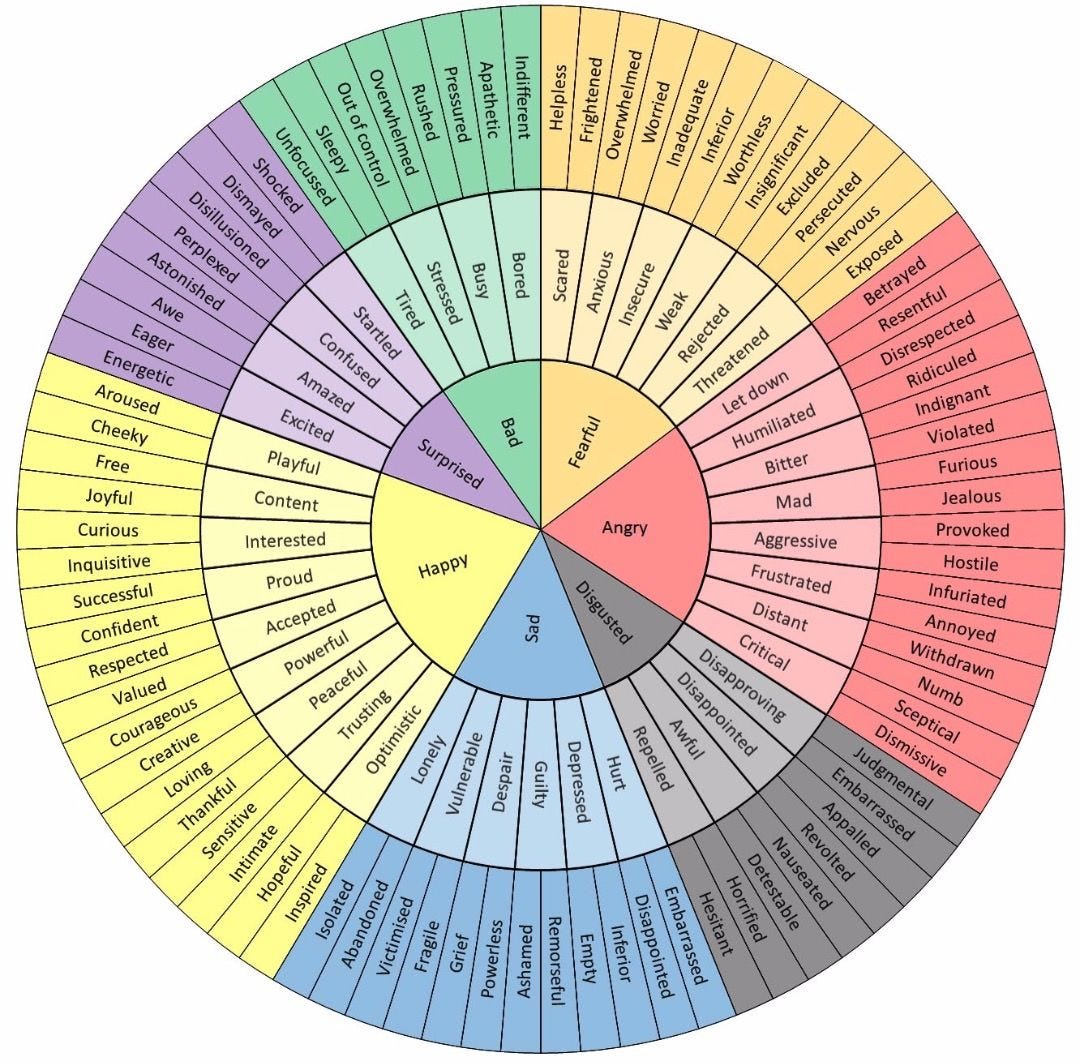}
    \caption{Wheel of emotions by \citet{Willcox1982}.}
    \label{app:fig:emotions}
\end{figure}

In this section, you can find the pilot results (Table~\ref{app:tab:alpha_pilot} shows the Kippendorff's $\alpha$ for the pilot annotation), the annotation guidelines used for the pilot and main annotation experiments (Table~\ref{app:tab:annot_guidelines}), as well as the feelings wheel (Figure~\ref{app:fig:emotions}).

\begin{table}[!t]
\centering
\small
\begin{tabular}{l l c}
\hline
\textbf{Category} & \textbf{Label} & \textbf{Krippendorff's $\alpha$} \\
\hline
          & Negative & 0.317 \\
Sentiment & Neutral  & 0.180 \\
        & Positive & 0.586 \\
\hline
 & Happy      & 0.451 \\
 & Bad        & 0.283 \\
 & Sad        & 0.281 \\
Emotion (Inner Ring) & Surprised  & 0.241 \\
 & Angry      & 0.191 \\
 & Fearful    & 0.121 \\
 & Disgusted  & 0.085 \\
\hline
\end{tabular}
\caption{Krippendorff's alpha for pilot annotation for sentiment labels and inner-ring emotion categories}
\label{app:tab:alpha_pilot}
\end{table}

\begin{table*}
\centering
\small
\renewcommand{\arraystretch}{1.3}
\begin{tabularx}{\textwidth}{>{\raggedright\arraybackslash}p{4cm} X}
\toprule
\textbf{Category} & \textbf{Guidelines} \\
\midrule
Task Overview & Label Greek proverbs with their overall sentiment (Positive, Neutral, Negative) and the emotion(s) expressed using the predefined list. Focus on the figurative meaning, provide consistent, evidence-based labels, record confidence, and note any normalization issues. \\
\midrule
Data Privacy & All data collected will be used solely for research and analysis. Personal information will not be collected or shared, ensuring compliance with data protection regulations (e.g., GDPR). \\
\midrule
Sentiment Labeling & Determine the overall sentiment expressed in the proverb. Multiple sentiments may be labeled if they co-occur or depending on interpretation. \\
\midrule
Emotion Labeling & Identify the strongest emotion(s) expressed. Multiple emotions can be labeled if they co-occur or depending on interpretation. \\
\midrule
Multiple Interpretations & If the proverb allows different interpretations leading to opposing emotions or sentiments, note these in the comments. \\
\midrule
Confidence Score & Record confidence (1--5) for both sentiment and emotion labeling (1 = very unsure, 5 = very confident). \\
\midrule
Missing Emotions & If the provided emotions do not fully capture the proverb's feeling, select the closest matching emotion and also note any new emotion in the “Other Emotion” field. \\
\midrule
Normalization Flag & If the normalized form of the proverb is incorrect, explain the issue in the normalization flag field. \\
\midrule
\textbf{Sentiment Definitions} & \\
Positive & Encouraging, motivating, or uplifting. Example: « \foreignlanguage{greek}{(Στη βράση κολλάει το σίδερο}.» → Urges action and initiative. \\
Neutral & Balanced, realistic, or advisory without strong praise or criticism. Example: « \foreignlanguage{greek}{(Αγάπα το φίλο σου με τα ελαττώματά του}.» → Advises acceptance without judgment. \\
Negative & Critical, pessimistic, or warning of harm or wrongdoing. Example: « \foreignlanguage{greek}{(Κακό σκυλί ψόφο δεν έχει.}» → Observes persistent harmful behavior. \\
Rule & Assign sentiment based on the overall metaphorical message or attitude, not individual literal words. \\
\midrule
\textbf{Emotion Guidelines} & \\
Rule & Select emotion(s) only from the approved list. Label based on overall feeling, not individual words. Multiple emotions may co-occur. Note opposing emotions in comments if interpretations differ. \\
Missing Emotion Handling & If an emotion is present but not listed, choose the closest match and record the new emotion in “Other Emotion”. \\
Confidence Score & Record confidence (1--5) for emotion labeling. \\
Approved Emotion List & Out of control, Overwhelmed, Rushed, Pressured, Apathetic, Indifferent, Helpless, Frightened, Worried, Inadequate, Inferior, Worthless, Insignificant, Excluded, Persecuted, Nervous, Exposed, Betrayed, Resentful, Disrespected, Ridiculed, Indignant, Violated, Furious, Jealous, Provoked, Hostile, Infuriated, Annoyed, Withdrawn, Numb, Sceptical, Dismissive, Judgmental, Embarrassed, Appalled, Revolted, Detestable, Horrified, Hesitant, Disappointed, Empty, Remorseful, Ashamed, Powerless, Grief, Fragile, Victimised, Abandoned, Isolated, Inspired, Hopeful, Intimate, Sensitive, Thankful, Loving, Creative, Courageous, Valued, Respected, Confident, Successful, Inquisitive, Curious, Joyful, Free, Cheeky, Aroused, Energetic, Eager, Awe, Astonished, Perplexed, Disillusioned, Dismayed, Shocked. \\
\bottomrule
\end{tabularx}
\caption{Annotation guidelines for sentiment and emotion in Greek proverbs.}\label{app:tab:annot_guidelines}
\end{table*}

\section{Annotator Confidence}\label{app:sec:confidence}
We also examine the confidence levels of the annotators per label for both sentiment and emotion task. The scale for confidence was from 1 to 5. The average confidence scores show that annotators were generally less confident when labeling neutral and negative sentiments, with `neutral' scoring 3.15 and `negative' 3.40, while `positive' labels received slightly higher confidence at 3.66. Looking at the more granular emotion labels, confidence was lowest for labels such as `empty' (2.09), `withdrawn' (2.61), and `fragile' (2.68), suggesting these emotions were more ambiguous or harder for annotators to assess. In contrast, annotators felt most confident about intense or clear-cut emotions such as `furious' (4.28), `revolted' (4.06), and `frightened' (4.00). Overall, this pattern indicates that subtle or internalized emotions tend to receive lower confidence, while strong or externally expressed emotions are easier to identify consistently.

\section{Prompts}\label{app:sec:prompts}

The prompts used for the model evaluation can be found in this section. Table shows the prompts for the zero-shot and probabilistic setting for sentiment, while Table shows the one for emotion. 

\begin{table*}[!t]
\centering
\small
\renewcommand{\arraystretch}{1.1}
\begin{tabularx}{\linewidth}{X}
\toprule
\textbf{Sentiment Analysis Prompts for Greek Proverbs} \\
\midrule

\textbf{Zero-shot Multi-label Sentiment Prompt.}
You are a sentiment analyst. Analyze the proverb \texttt{"\{proverb\}"}.
Identify all applicable sentiments (positive, negative, neutral).
Multiple labels are allowed; briefly justify each.\\
\textbf{Output:}
\texttt{Sentiments: <labels>, Justification: <text>} \\

\midrule

\textbf{Probabilistic Few-shot Sentiment Prompt.}
Assign sentiment percentages (0--100) summing to exactly 100.\\
\textbf{Examples:} \parbox[t]{\linewidth}{\texttt{
E1: \foreignlanguage{greek}{Θηβαίοι κι Αθηναίοι και κακοί Μυτιληναίοι \\άλλα λένε το βράδι κι άλλα κάνουν το πρωΐ.} → Pos 0\%, Neg 100\%, Neu 0\%\\
E2: \foreignlanguage{greek}{Τσείνες π΄ μ΄ αγαπά, με κάνει τσαι κλαίου,\\τσαι τσείνες π΄ με μ΄σά,
με κάν΄ τσαι γελού
} → Pos 16.67\%, Neg 25\%, Neu 58.33\%\\
E3: \foreignlanguage{greek}{Δούλευε και δουλειά να μη σε λείπ΄} → Pos 41.67\%, Neg 16.67\%, Neu 41.67\%
}}\\
Now analyze the proverb \texttt{"\{proverb\}"} and assign percentages to each sentiment.\\
\textbf{Output:}
\texttt{Positive, Negative, Neutral (\%) + Justification} \\

\bottomrule
\end{tabularx}
\caption{Sentiment annotation prompts.}\label{app:tab:prompts_sentiment}
\end{table*}

\begin{table*}[!t]
\centering
\small
\renewcommand{\arraystretch}{1.1}
\begin{tabularx}{\linewidth}{X}
\toprule
\textbf{Emotion Analysis Prompts for Greek Proverbs} \\
\midrule

\textbf{Zero-shot Multi-label Emotion Prompt.}
You are an emotion analyst. Analyze the proverb \texttt{"\{proverb\}"}.
Identify all applicable emotions (happy, angry, bad, fearful, sad, disgusted, surprise).
Multiple labels are allowed; briefly justify each.\\
\textbf{Output:}
\texttt{Emotions: <labels>, Justification: <text>} \\

\midrule

\textbf{Probabilistic Few-shot Emotion Prompt.}
Assign emotion percentages (0--100) summing to exactly 100.\\
\textbf{Examples:} \parbox[t]{\linewidth}{\texttt{
E1: \foreignlanguage{greek}{Άε να κορεύεσαι} → angry 73\%, disgusted 18\%\\
E2: \foreignlanguage{greek}{Άμον ντ εγόρασα πουλώ σ' άτο} → happy 36\%, bad 27\%, angry 18\%\\
E3: \foreignlanguage{greek}{Άνοιξε τα μάτια σ' τέσσερα} → fearful 27\%\\
E4: \foreignlanguage{greek}{Έμαθε και γένεται, τώρα δεν ξεγένεται} → sad 36\%\\
E5: \foreignlanguage{greek}{Είπε ν' αγιάση κι' εξεπάγιασε} → surprise 27\%
}}\\
Now analyze the proverb \texttt{"\{proverb\}"} and assign percentages to each emotion.\\
\textbf{Output:}
\texttt{happy, angry, bad, fearful, sad, disgusted, surprised (\%)} \\

\bottomrule
\end{tabularx}
\caption{Emotion prompts.}\label{app:tab:prompts_emotion}
\end{table*}




\section{More on Error Analysis}\label{app:sec:error_analysis}
We delve further into examining the trustworthiness of the models used in this study by calculating Precision, Recall, and F1 score per label for both sentiment and emotion. The results can be found in Tables~\ref{app:tab:sentiment_zero_perlabel}, \ref{app:tab:sentiment_prob_perlabel}, \ref{app:tab:emotion_zero_perlabel}, \ref{app:tab:emotion_prob_perlabel}. In addition, we examine use the probabilistic setting as an additional error analysis setting by calculating the Mean Squared Error (MSE) for the two tasks. The results can be found in Tables~\ref{app:tab:mse_results_sentiment} and~\ref{app:tab:mse_results_emotion}.

Looking at sentiment, the MSE values confirms our main result analysis. When it comes to overall performance, GPT-5mini (0.23) and Llama3-70B (0.25) are the best overall, while Krikri-8B (0.31) and Mistral-7B (0.33) have higher overall errors, so their predictions are less accurate. The pattern is the same for the positive label, however, for negative sentiment Krikri-8B exceeds (0.13) the other models but struggles with the neutral label (0.54) compared to the others. All in all, GPT-5mini is the most balanced overall as it has the lowest overall MSE and handles all sentiment types reasonably well. Llama3-70B is close in performance to GPT-5mini but slightly worse overall. Krikri-8B is good at negative sentiment but poor at neutral, while Mistral-7B has relatively high errors across the board, especially for positive sentiment.

For emotion, Krikri-8B has a global MSE of 0.22, which is slightly higher than Mistral-B (0.21) and much higher than GPT-5mini (0.15) and Llama3-70B (0.14), which indicates that the high F1 scores of the first two models can be misleading. We observe quite the variation when it comes to label-wise errors per model. Mistral-7B predicts `Happy' best (0.32), while GPT-5mini is worse (0.25). With  `Surprised', GPT-5mini has the lowest error (0.09), and Krikri-8B the highest (0.46). The best model predicting `Angry' is Llama3-70B.


\begin{table*}[!t]
\centering
\small
\renewcommand{\arraystretch}{1.1}
\begin{tabular}{lcccccccccccc}
\toprule
 & \multicolumn{3}{c}{\textbf{Krikri-8B}} &
   \multicolumn{3}{c}{\textbf{Mistral-7B}} &
   \multicolumn{3}{c}{\textbf{GPT-5mini}} &
   \multicolumn{3}{c}{\textbf{Llama3-70B}}\\
\cmidrule(lr){2-4} \cmidrule(lr){5-7} \cmidrule(lr){8-10} \cmidrule(lr){11-13}
\textbf{Label}
& P & R & F1 
& P & R & F1 
& P & R & F1 
& P & R & F1 \\
\midrule
Negative & 0.80 & 0.51 & 0.62 &  0.81 & 0.57 & 0.66 & 0.88 & 0.80 & 0.84 & 0.84 & 0.88 & 0.86\\
Neutral  & 0.63 & 0.55 & 0.59 &  0.66 & 0.67 & 0.66 & 0.69 & 0.73 & 0.71 & 0.68 & 0.72 & 0.70\\
Positive & 0.36 & 0.55 & 0.43 & 0.39 & 0.55 & 0.45 & 0.69 & 0.62 & 0.65 & 0.64 & 0.45 & 0.52 \\
\bottomrule
\end{tabular}
\caption{Zero-shot sentiment per-label performance across models (P = Precision, R = Recall, F1 = F1 score).}\label{app:tab:sentiment_zero_perlabel}
\end{table*}

\begin{table*}[!t]
\centering
\small
\renewcommand{\arraystretch}{1.1}
\begin{tabular}{lcccccccccccc}
\toprule
 & \multicolumn{3}{c}{\textbf{Krikri-8B}} &
   \multicolumn{3}{c}{\textbf{Mistral-7B}} &
   \multicolumn{3}{c}{\textbf{GPT-5mini}} &
   \multicolumn{3}{c}{\textbf{Llama3-70B}}\\
\cmidrule(lr){2-4} \cmidrule(lr){5-7} \cmidrule(lr){8-10} \cmidrule(lr){11-13}
\textbf{Label}
& P & R & F1 
& P & R & F1 
& P & R & F1 
& P & R & F1 \\
\midrule
Negative & 0.85 & 0.86 & 0.86 & 0.85 & 0.68 & 0.75 & 0.91 & 0.73 & 0.81 & 0.95 & 0.55 & 0.69 \\
Neutral  & 0.61 & 0.56 & 0.59 & 0.68 & 0.31 & 0.43 & 0.67 & 0.77 & 0.72 & 0.64 & 0.57 & 0.61 \\
Positive & 1.00 & 0.01 & 0.03 & 0.46 & 0.22 & 0.30 & 0.55 & 0.66 & 0.60 & 0.79 & 0.44 & 0.57 \\

\bottomrule
\end{tabular}
\caption{Probabilistic few-shot sentiment per-label performance across models (P = Precision, R = Recall, F1 = F1 score).}\label{app:tab:sentiment_prob_perlabel}
\end{table*}

\begin{table*}[ht]
\centering
\small
\renewcommand{\arraystretch}{1.1}
\begin{tabular}{lcccccccccccc}
\toprule
 & \multicolumn{3}{c}{\textbf{Krikri-8B}} &
   \multicolumn{3}{c}{\textbf{Mistral-7B}} &
   \multicolumn{3}{c}{\textbf{GPT-5mini}}&
   \multicolumn{3}{c}{\textbf{Llama3-70B}} \\
\cmidrule(lr){2-4} \cmidrule(lr){5-7} \cmidrule(lr){8-10} \cmidrule(lr){11-13}
\textbf{Label} & P & R & F1 & P & R & F1 & P & R & F1 &  P & R & F1\\
\midrule
Angry      & 0.59 & 0.69 & 0.63  & 0.60 & 0.37 & 0.46 & 0.77 & 0.35 & 0.48 & 0.58 & 0.78 & 0.66 \\
Happy      & 0.41 & 0.94 & 0.57  & 0.52 & 0.50 & 0.51 & 0.77 & 0.40 & 0.53 & 0.72 & 0.19 & 0.30 \\
Sad        & 0.14 & 0.90 & 0.24  & 0.11 & 0.57 & 0.19 & 0.20 & 0.63 & 0.30 & 0.12 & 0.80 & 0.21 \\
Fearful    & 0.25 & 0.21 & 0.23  & 0.20 & 0.34 & 0.25 & 0.24 & 0.30 & 0.26 & 0.17 & 0.15 & 0.15 \\
Bad        & 0.09 & 0.41 & 0.15  & 0.20 & 0.03 & 0.06 & 0.12 & 0.48 & 0.19 & 0.07 & 0.07 & 0.07 \\
Surprised  & 0.12 & 0.17 & 0.14  & 0.18 & 0.47 & 0.26 & 0.20 & 0.10 & 0.13 & 0.24 & 0.13 & 0.17 \\
Disgusted  & 0.05 & 0.09 & 0.06  & 0.50 & 0.04 & 0.08 & 0.10 & 0.50 & 0.17 & 0.08 & 0.77 & 0.15 \\
\bottomrule
\end{tabular}
\caption{Zero-shot emotion per-label performance across models (P = Precision, R = Recall, F1 = F1-score).}\label{app:tab:emotion_zero_perlabel}
\end{table*}

\begin{table*}[ht]
\centering
\small
\renewcommand{\arraystretch}{1.1}
\begin{tabular}{lcccccccccccc}
\toprule
 & \multicolumn{3}{c}{\textbf{Krikri-8B}} &
   \multicolumn{3}{c}{\textbf{Mistral-7B}} &
   \multicolumn{3}{c}{\textbf{GPT-5mini}}&
   \multicolumn{3}{c}{\textbf{Llama3-70B}} \\
\cmidrule(lr){2-4} \cmidrule(lr){5-7} \cmidrule(lr){8-10} \cmidrule(lr){11-13}
\textbf{Label} & P & R & F1 & P & R & F1 & P & R & F1 & P & R & F1 \\
\midrule
Angry      & 0.59 & 0.35 & 0.44 & 0.48 & 0.22 & 0.30 & 0.67 & 0.37 & 0.47 & 0.54 & 0.97 & 0.70 \\
Happy      & 0.56 & 0.33 & 0.42 & 0.46 & 0.24 & 0.32 & 0.64 & 0.54 & 0.59 & 0.86 & 0.18 & 0.30 \\
Sad        & 1.00 & 0.07 & 0.12 & 0.11 & 0.10 & 0.10 & 0.27 & 0.36 & 0.31 & 0.27 & 0.10 & 0.14 \\
Fearful    & 1.00 & 0.02 & 0.04 & 0.33 & 0.09 & 0.14 & 0.27 & 0.17 & 0.20 & 0.40 & 0.04 & 0.07 \\
Bad        & 0.00 & 0.00 & 0.00 & 0.06 & 0.19 & 0.09 & 0.11 & 0.33 & 0.16 & 0.02 & 0.07 & 0.03 \\
Surprised  & 0.12 & 0.66 & 0.20 & 0.09 & 0.36 & 0.15 & 0.36 & 0.13 & 0.20 & 0.15 & 0.06 & 0.09 \\
Disgusted  & 0.00 & 0.00 & 0.00 & 0.00 & 0.00 & 0.00 & 0.12 & 0.23 & 0.16 & 0.50 & 0.05 & 0.08  \\
\bottomrule
\end{tabular}
\caption{Probabilistic emotion per-label performance across models (P = Precision, R = Recall, F1 = F1-score).}\label{app:tab:emotion_prob_perlabel}
\end{table*}

\begin{table}[!t]
\centering
\small
\begin{tabular}{lcccc}
\toprule
\textbf{Model} &
\textbf{MSE} &
\textbf{MSE$_{pos}$} &
\textbf{MSE$_{neg}$} &
\textbf{MSE$_{neu}$} \\
\midrule
Krikri-8B & 0.31 & 0.17 & 0.21 & 0.54 \\
Mistral-7B & 0.33 & 0.26 & 0.32 & 0.41 \\
GPT-5mini & 0.23 & 0.13 & 0.28 & 0.28 \\
Llama3-70B & 0.25 & 0.16 & 0.29 & 0.31 \\
\bottomrule
\end{tabular}
\caption{Mean Squared Error (MSE) between predicted sentiment probabilities and gold labels. Lower is better.}
\label{app:tab:mse_results_sentiment}
\end{table}

\begin{table*}[!t]
\centering
\small
\begin{tabular}{lcccccccc}
\toprule
\textbf{Model} & \textbf{Global} & \textbf{Disgusted} & \textbf{Sad} & \textbf{Bad} & \textbf{Fearful} & \textbf{Happy} & \textbf{Angry} & \textbf{Surprised} \\
\midrule
Krikri-8B & 0.22 & 0.07 & 0.09 & 0.10 & 0.15 & 0.29 & 0.40 & 0.46 \\
Mistral-7B & 0.21 & 0.07 & 0.12 & 0.15 & 0.15 & 0.32 & 0.38 & 0.32 \\
GPT-5mini & 0.15 & 0.08 & 0.09 & 0.11 & 0.14 & 0.25 & 0.32 & 0.09 \\
Llama3-70B & 0.14 & 0.07 & 0.09 & 0.10 & 0.14 & 0.23 & 0.30 & 0.10 \\
\bottomrule
\end{tabular}
\caption{Mean Squared Error (MSE) between gold emotion labels and predicted probabilities for multiple models. Lower is better.}
\label{app:tab:mse_results_emotion}
\end{table*}

\section{Models and Budget}\label{app:sec:models_budget}
The open-sourced models were used in their quantized forms through hugging face, with only prompt configurations adjusted to optimize task performance.
Llama 70B and GPT-4o were accessed respectively through AWS Bedrock and the OpenAI API.
We used fixed inference settings, such as, temperature=0.7, topP=0.9.
The budget reflected the usage costs of the APIs charged by each provider (AWS and OpenAI), with no additional expenses for model training or fine-tuning.

\end{document}